\documentclass[a4paper,11pt]{article}

\usepackage[T1]{fontenc}
\usepackage{epsfig}
\usepackage{calc}
\usepackage[ruled,vlined]{algorithm2e}
\usepackage{amssymb}
\usepackage{amstext}
\usepackage{amsmath}
\usepackage{amsthm}
\usepackage{pslatex}
\usepackage{apalike}
\usepackage{booktabs}
\usepackage{array}
\usepackage{tikz}
\usepackage[hidelinks]{hyperref}
\usepackage{xcolor}
\usepackage[bottom]{footmisc}
\usepackage{authorversion}
\setAVvenue{ICAART 2027}
\setAVnotice{Author's accepted version of a paper accepted at the 19th International Conference on Agents and Artificial
Intelligence (ICAART 2027), Valletta, Malta, 23--26 February 2027. The final version is published by SCITEPRESS:
DOI to be added. This version does not use the publisher's layout.}

\hypersetup{pdfkeywords={LLM Agents, LLM-as-a-Judge, Evaluation Reliability, Measurement Error, Randomized Auditing, Partial Identification, Confidence Sequences, Leaderboard Inference},
  pdftitle={Transport-Safe Evaluation of LLM Agents: Trajectory-Dependent Judge Error and Anytime-Valid Auditing},
  pdfauthor={Yulliwas Ameur, Samia Bouzefrane, Soumya Banerjee},pdfsubject={Author's accepted version, ICAART 2027, paper 38}}

\newtheorem{theorem}{Theorem}
\newtheorem{proposition}{Proposition}

\newcommand{\Yhat}{\widehat{Y}}
\newcommand{\Prob}{\mathbb{P}}
\newcommand{\E}{\mathbb{E}}
\newcommand{\figref}[1]{Figure~\ref{#1}}

\SetAlFnt{\footnotesize}\SetAlgoCaptionLayout{centerline}

\begin{document}

\title{Transport-Safe Evaluation of LLM Agents: Trajectory-Dependent Judge Error and Anytime-Valid Auditing}

\author{\authorname{Yulliwas Ameur\sup{1,2}, Samia Bouzefrane\sup{2} and Soumya Banerjee\sup{2}}
\affiliation{\sup{1}Efrei Research Lab, Universit\'e Paris-Panth\'eon-Assas, Villejuif, France}
\affiliation{\sup{2}CEDRIC, Conservatoire National des Arts et M\'etiers (Cnam), Paris, France}
\email{yulliwas.ameur@efrei.fr, samia.bouzefrane@lecnam.net, dr.soumya@ieee.org}
}


\abstract{Large language model (LLM) agents are often ranked by an LLM judge whose error rates are measured once and reused for every agent. The corrected success rates can then be biased when the judge's errors depend on the trajectory and agents differ in the trajectories they produce for successes or failures. We derive the exact asymptotic bias and give a chain of checks from identification to simultaneous ranks. On a four-agent semi-synthetic benchmark, a scalar correction with oracle calibration has maximum bias 0.245 and mean pointwise coverage 0.016, and never recovers the full ranking (0/250 replications). Augmented inverse-probability weighting (AIPW) with a uniform or a design-aware random audit keeps the maximum bias below 0.003, with mean pointwise coverage of 0.939 and 0.950. In a 48-cell stress test, effective target support rises only from 0.209 to 0.805 as calibration grows from 600 to 4,800 labels, while audited AIPW keeps a mean absolute error of 0.018--0.020. Without calibration support, or with judge dependence within an externally justified bound, we report valid bounds or sensitivity sets and carry them into rank sets. An \mbox{e-process} confidence sequence stays valid, for a stable target, under predictable adaptive auditing and arbitrary stopping.}

\maketitle

% ---------------------------------------------------------------------------------------------
\section{\uppercase{Introduction}}
\label{sec:intro}

Agents built on large language models (LLMs) solve tasks through trajectories of actions, tool calls and outputs.\footnote{\label{fn:ai}Text of this section drafted or edited with Claude \cite{anthropic2026}; see the Acknowledgements.} Expert labels are expensive, so many evaluations use a second LLM as a \emph{judge}: its sensitivity and specificity are estimated once on a hand-labeled sample, and every agent's judged success rate is corrected with these two numbers.

This correction assumes that the judge's error rates are the same for every agent. A judge, however, reads trajectories, not agent names, and its errors can change with trajectory length, task family, number of tool calls, or the kind of failure. Different agents produce different trajectories, so error rates averaged over the calibration trajectories need not hold for another agent, even when the judge never sees which agent it grades.

\textbf{A concrete warning sign.} AgentRewardBench provides expert labels and automatic judge verdicts for web-agent trajectories \cite{lu2025}. A prior analysis of these public outputs, not independently rerun for this paper, reports judge sensitivity falling from 0.840/0.884/0.876 in the shortest trajectory tercile to 0.182/0.364/0.477 in the long or capped tercile, for the three judges labeled AER, GPT-4o and GPT-4o-mini in that benchmark. Leave-one-agent-out scalar calibration then gives reported biases from $+0.097$ to $-0.186$. These figures motivate the problem; they are not new evidence and do not show that length alone causes the drop.

\textbf{Question and scope.} When can a judge-based leaderboard be trusted, and what should be reported when it cannot? Our contribution is integrative: augmented inverse-probability weighting (AIPW), latent classes, confidence sequences and rank inference are not new; we connect them through the agent-specific target and state each conclusion's assumptions.

\textbf{Contributions.}
(1)~The exact limit of a transported scalar correction and when it reverses a ranking (Section~\ref{sec:transport}); (2)~one- and three-judge identification, bounds and sensitivity sets (Section~\ref{sec:ident}); (3)~randomized auditing with cross-fitted AIPW, bounded allocation and an anytime-valid e-process (Section~\ref{sec:audit}); (4)~rank sets with sampling, identification and cluster uncertainty (Section~\ref{sec:ranks}); proof sketches (Section~\ref{sec:proofs}) and baseline comparisons (Section~\ref{sec:evidence}). Artificial intelligence (AI) assistance is disclosed in the Acknowledgements.

% ---------------------------------------------------------------------------------------------
\section{\uppercase{Related Work}}
\label{sec:related}

\textbf{LLM judges and agent evaluation.} LLM judges scale evaluation but show position, verbosity, self-preference, prompt and task-complexity effects \cite{zheng2023,gu2026,thakur2025,bavaresco2025}.\textsuperscript{\ref{fn:ai}} Agent benchmarks add long horizons and tool use \cite{jimenez2024,kapoor2025,lu2025}; recent work corrects noisy judge scores with sensitivity and specificity or with semiparametric estimators \cite{lee2025,chen2026b}, and calibration instability can reverse comparisons \cite{fiedler2026}.

\textbf{Misclassification and latent classes.} Error rates that vary across covariates can invalidate a transported scalar correction \cite{edwards2023}. Missing support calls for partial identification \cite{richardson2014,obradovic2024}; dependent LLM panels can be unidentified without external restrictions \cite{chen2026a}. Identification without a gold standard from conditionally independent tests is classical \cite{dawid1979,allman2009,zhang2016}; covariates need extra conditions \cite{ouyang2022}, and hidden dependence can bias latent-class estimates \cite{albert2004,keddie2023}.

\textbf{Two-phase and rank inference.} AIPW and optimal validation sampling are well established \cite{robins1994,hahn1998,tao2020}. Concurrent work studies two-stage human review \cite{kim2026}, best-arm identification with judges \cite{ao2026}, and e-process monitoring \cite{zhou2026}. Confidence sequences allow optional stopping \cite{howard2021,waudbysmith2024}; rank inference needs simultaneous uncertainty \cite{hall2009,klein2020,mogstad2024,almohamad2021}.

\textbf{Closest baselines.} Our scalar baseline is the confusion-matrix plug-in \cite{lee2025} with oracle calibration, so its remaining error is transport error. Our uniform-audit baseline is the efficient-influence-function (EIF) and prediction-powered inference (PPI) route \cite{chen2026b}. We add the trajectory-conditional target, support and dependence diagnostics, design-aware allocation, anytime monitoring and rank propagation.

% ---------------------------------------------------------------------------------------------
\section{\uppercase{Scalar Transport}}
\label{sec:transport}

An episode of agent $a$ produces $(X,Y,\Yhat)$: trajectory covariates $X$, expert success $Y\in\{0,1\}$ and judge verdict $\Yhat\in\{0,1\}$.\textsuperscript{\ref{fn:ai}} Agent $a$ induces a law $\pi_a$ and has ability $\theta_a=\Prob_a(Y=1)\in(0,1)$. We assume throughout that the judge is \emph{blind given $(X,Y)$}: $\Prob_a(\Yhat=1\mid X,Y)$ is the same for every agent and for the calibration law. Its trajectory-dependent sensitivity and specificity are
\begin{equation}
\begin{aligned}s(x)&=\Prob(\Yhat=1\mid Y=1,X=x),\\ t(x)&=\Prob(\Yhat=0\mid Y=0,X=x),\end{aligned}
\label{eq:st}
\end{equation}
with Youden index $J(x)=s(x)+t(x)-1$. A calibration law $\pi_0$ with $\Prob_0(Y=1)\in(0,1)$ gives scalar averages $s_0=\E_0\{s(X)\mid Y=1\}$ and $t_0=\E_0\{t(X)\mid Y=0\}$. The transported Rogan--Gladen estimator \cite{rogan1978} is $\widehat\theta_a^{RG}=(\widehat p_a+\widehat t_0-1)/(\widehat s_0+\widehat t_0-1)$, where $\widehat p_a$ is the judge-positive rate of agent $a$. Let $s^\star_a=\E_a\{s(X)\mid Y=1\}$ and $t^\star_a=\E_a\{t(X)\mid Y=0\}$.

\begin{theorem}[Transport bias]\label{thm:transport}
Assume judge blindness given $(X,Y)$, consistent estimates $\widehat s_0\to s_0$, $\widehat t_0\to t_0$, $\widehat p_a\to\Prob_a(\Yhat=1)$, and $J_0=s_0+t_0-1\neq0$. Then $\widehat\theta_a^{RG}\to\theta_a+B_a$ in probability, with
\begin{equation}
B_a=\frac{\theta_a(s^\star_a-s_0)+(1-\theta_a)(t_0-t^\star_a)}{J_0}.
\label{eq:bias}
\end{equation}
The correction is consistent for agent $a$ exactly when the numerator of \eqref{eq:bias} vanishes.
\end{theorem}

If $w_{a,1}=d\pi_a(\cdot\mid Y=1)/d\pi_0(\cdot\mid Y=1)$ exists, then $s^\star_a-s_0=\mathrm{Cov}_0\{w_{a,1}(X),s(X)\mid Y=1\}$, and similarly for specificity: the bias couples trajectory shift with judge accuracy, not just a low average accuracy. For two agents with $\theta_a>\theta_b$, the transported order reverses exactly when $B_b-B_a>\theta_a-\theta_b$. More judge compute is not a generic repair: a majority of an odd number $n\geq3$ of calls, independent and identically distributed given $(X,Y)$, pushes each trajectory-level accuracy other than $0$, $1/2$ and $1$ further from $1/2$, so the bias can grow or shrink.

% ---------------------------------------------------------------------------------------------
\section{\uppercase{Identification}}
\label{sec:ident}

\subsection{One Judge and Calibration Support}

Calibration records $(X,Y,\Yhat)$ identify $s(x)$ and $t(x)$ only where both classes are observed.\textsuperscript{\ref{fn:ai}} Target records identify $\mu_a(x)=\Prob_a(\Yhat=1\mid X=x)$. Let $\mathcal O$ be the set of $x$ where $s(x)$ and $t(x)$ are both identified and $J(x)\neq0$, and split its complement into $\mathcal U_0$ (neither curve identified, or both identified with $J(x)=0$) and $\mathcal U_1$ (exactly one curve identified).

\begin{theorem}[One-judge identification]\label{thm:onejudge}
Assume judge blindness given $(X,Y)$ and $\Prob_a(X\in\mathcal O)=1$, i.e., calibration observes both classes wherever the target puts mass and the judge is informative there. Then
\begin{equation}
\theta_a=\E_a\!\left[\frac{\mu_a(X)+t(X)-1}{J(X)}\right].
\label{eq:onejudge}
\end{equation}
Without this condition, if $\Prob_a(X\in\mathcal U_0)>0$, $\theta_a$ is not nonparametrically point identified from the calibration law and agent~$a$'s own records.
\end{theorem}

For estimation, $|J|\geq j_{\min}>0$ keeps the weights in \eqref{eq:onejudge} bounded; near-zero calibration support gives weak identification and high variance. Let $\mathcal U=\mathcal U_0\cup\mathcal U_1$, $\rho_a=\Prob_a(X\in\mathcal U)$ and $\nu_a=\E_a[\mathbf 1\{X\in\mathcal O\}\{\mu_a(X)+t(X)-1\}/J(X)]$.

\begin{proposition}[Support-failure set]\label{prop:sharp}
Under judge blindness and no further restriction on $\mathcal U$, $\theta_a\in\Theta_a=[\nu_a,\nu_a+\rho_a]$. This set is sharp up to its endpoints, given the calibration law and agent~$a$'s own records, when $\Prob_a(X\in\mathcal U_1)=0$; otherwise, or when other agents' records are pooled through the shared curves, it is a valid outer set. If $\ell(x)\leq\Prob_a(Y=1\mid X=x)\leq u(x)$ on $\mathcal U$, the endpoints become $\nu_a+\E_a[\mathbf 1\{X\in\mathcal U\}\ell(X)]$ and $\nu_a+\E_a[\mathbf 1\{X\in\mathcal U\}u(X)]$.
\end{proposition}

In an executed four-stratum example, $\nu_a=0.445$ and $\rho_a=0.20$, so $\Theta_a=[0.445,0.645]$; the truth 0.525 lies inside, whereas naive extrapolation from short trajectories gives 0.431, outside the set.

\textbf{A rich covariate is data-hungry.} We take $X=(\text{length},\ \text{task family},\ \text{tool-call intensity})$ with $4\times4\times3=48$ cells and require at least five calibration labels per outcome class in a cell. The mean target mass with such support is only 0.209, 0.421, 0.643 and 0.805 for 600, 1,200, 2,400 and 4,800 calibration labels. A cellwise plug-in with pooled fallback has mean absolute error (MAE) 0.188, 0.158, 0.119 and 0.077, whereas AIPW with a 12\% uniform randomized audit stays at 0.018--0.020. 

\subsection{Three Judges and Dependence}

For three binary judges, at each $x$, let $\mu_j$, $c_{jk}$ and $T$ be the first, pairwise central and third central moments given $X=x$.

\begin{theorem}[Three-judge model-based inversion]\label{thm:three}
If the judges are conditionally independent given $(X,Y)$, each $J_j(x)>0$, and $\theta(x)=\Prob_a(Y=1\mid X=x)\in(0,1)$ for $\pi_a$-almost every $x$, then, with moments under $\Prob_a$ and $v=\theta(1-\theta)$,
\begin{equation}
c_{jk}=vJ_jJ_k,\qquad T=v(1-2\theta)J_1J_2J_3,
\label{eq:three}
\end{equation}
and these moments identify $\theta(x)$ and all operating curves at $x$, hence $\theta_a=\E_a\theta(X)$. Two judges are generically insufficient.
\end{theorem}

Conditional independence can fail: the public-data source analysis (source-preserved, not rerun) reports within-$(Y,\text{length})$ dependence $\psi=0.41$--$0.70$, including two judges that share a backbone. Let $F(\theta,J)$ be the four moments $(c_{12},c_{13},c_{23},T)$ implied by independence at $x$, $M$ the observed moments, and $\Gamma$ a bound on the componentwise deviations $|M-F(\theta(x),J(x))|$ at the true values. We define the dependence-envelope set
\begin{equation}
I_\Gamma=\{\theta\in[0,1]:\exists J,\ |M-F(\theta,J)|\leq\Gamma\},
\label{eq:envelope}
\end{equation}
where $J$ ranges over feasible values: $J_j\in(0,1]$, $t_j=1-\mu_j+\theta J_j\in[0,1]$ and $s_j=1-t_j+J_j\in[0,1]$, for a candidate value $\theta$ of $\Prob_a(Y=1\mid X=x)$. The set $I_\Gamma$ is a valid outer set relative to the declared envelope when every $J_j(x)>0$ (Section~\ref{sec:proofs}), and, if the envelope holds at $\pi_a$-almost every $x$, an outer set for $\theta_a$ is $\{\E_a\theta(X):\theta(x)\in I_\Gamma(x)\}$. The bound $\Gamma$ is \emph{not} identified from the same three-judge law: it must come from external human calibration, held-out perturbations, or a preregistered range; otherwise only the independence point and a sensitivity curve should be reported. In an executed shared-factor example, the independence point 0.502 misses the truth 0.525; a half-envelope gives $[0.457,0.547]$ and the simulation-matched oracle envelope $[0.413,0.597]$. As conditional dependence grows to 0.231, label-free MAE rises from 0.0085 to 0.0343, while AIPW with a 10\% randomized human audit stays between 0.0062 and 0.0096.

% ---------------------------------------------------------------------------------------------
\section{\uppercase{Randomized Human Auditing}}
\label{sec:audit}

The three-judge route needs no labels but depends on a model.\textsuperscript{\ref{fn:ai}} Randomized human auditing is less novel but stronger by design: its validity follows from a design the evaluator controls, so, when a human budget is available, we recommend it as the default and the judge panel as a sensitivity analysis.

\subsection{Cross-Fitted AIPW}

Every episode has phase-1 variables $V=(X,\Yhat)$. A human label is observed when $R=1$, with known design probability $\varphi(V)\in[\varphi_{\min},1]$. Let $m(v)=\E(Y\mid V=v)$. For episode $i$, fit $\widehat m_{-k(i)}$ without its fold $k(i)$ (folds fixed independently of $R_{1:n}$) and compute
\begin{equation}
\begin{split}\widehat\theta_{CF}=\frac1n\sum_{i=1}^n\Big[&\widehat m_{-k(i)}(V_i)\\&+\frac{R_i}{\varphi(V_i)}\{Y_i-\widehat m_{-k(i)}(V_i)\}\Big].\end{split}
\label{eq:aipw}
\end{equation}
If the audit indicators are drawn independently, with $\Prob(R_i=1\mid V_{1:n},Y_{1:n},R_{-i})=\varphi(V_i)$ and $\varphi$ fixed before auditing, this estimator is exactly design-unbiased for the finite-population mean $n^{-1}\sum_iY_i$, hence unbiased for $\theta_a$ when episodes are drawn independently from $\pi_a$, even when the imputation model is wrong (Section~\ref{sec:proofs}). In two waves, each episode belongs to one wave, fixed before any audit is drawn and independently of $Y_{1:n}$, and $\varphi(V_i)$ is its wave's audit rule evaluated at $V_i$. Second-wave summands are unbiased given the pilot if second-wave audits are drawn independently given it; pilot summands stay unbiased if their imputations use no second-wave audit data. Under positivity and $L_2$ consistency of $\widehat m$, its influence function is $m(V)-\theta+R\{Y-m(V)\}/\varphi(V)$, and it attains the semiparametric efficiency bound for the given design in the model that leaves the law of $(V,Y)$ unrestricted \cite{robins1994,hahn1998}.

\subsection{Positivity-Constrained Allocation}

Write $\sigma^2(v)=m(v)\{1-m(v)\}$. Minimizing the asymptotic variance under a mean audit budget gives the bounded Karush--Kuhn--Tucker (KKT) rule
\begin{equation}
\varphi^\star(v)=\min\{1,\max\{\varphi_{\min},\lambda^\star\sigma(v)\}\},
\label{eq:kkt}
\end{equation}
where $\lambda^\star$ meets any budget from $\varphi_{\min}$ up to $\varphi_{\min}+(1-\varphi_{\min})\Prob\{\sigma(V)>0\}$ (attained when $\sigma$ is bounded away from $0$ on $\{\sigma>0\}$, e.g., finitely many strata); in the interior it is Neyman allocation, $\varphi\propto\sigma$. Since $\sigma$ is unknown, a randomized pilot wave estimates it and a second wave applies \eqref{eq:kkt}. With finitely many strata, the two-wave rule converges to the oracle allocation as the pilot grows while its fraction vanishes. Over 500 replications, the variance ratio to the oracle falls from 1.076 at 20,000 episodes to 1.038 at $10^5$ and 1.018 at $5\times10^5$, while uniform auditing stays at 1.077.

\subsection{Anytime-Valid Monitoring}

Audits are often inspected repeatedly and stopped when an interval looks decisive. Fixed-time Wald intervals lose their coverage when they are recomputed and checked at every look. Let $\mathcal F_t=\sigma(V_i,R_i,R_iY_i:i\leq t)$ and
\begin{equation}
\begin{split}W_t(\theta)={}&\widehat m_{t-1}(V_t)-\theta\\&+\frac{R_t}{\varphi_t(V_t)}\{Y_t-\widehat m_{t-1}(V_t)\},\end{split}
\label{eq:W}
\end{equation}
where $\widehat m_{t-1}(\cdot)\in[0,1]$ and the audit rule $\varphi_t(\cdot)$ are chosen from $\mathcal F_{t-1}$ and evaluated at the new $V_t$, and $\Prob(R_t=1\mid\mathcal F_{t-1},V_t,Y_t)=\varphi_t(V_t)$. For a finite grid $\Lambda\subset(0,\varphi_{\min})$ fixed before monitoring, define the hedged e-process
\begin{equation}
\begin{split}E_t(\theta)=\frac{1}{2|\Lambda|}\sum_{\lambda\in\Lambda}\Big[&\prod_{i\leq t}\{1+\lambda W_i(\theta)\}\\&+\prod_{i\leq t}\{1-\lambda W_i(\theta)\}\Big],\end{split}
\label{eq:eprocess}
\end{equation}
and $C_t=\{\theta\in[0,1]:E_t(\theta)<1/\alpha\}$ \cite{howard2021,waudbysmith2024}.

\begin{theorem}[Anytime-valid audit]\label{thm:anytime}
Assume design randomization as above, past-measurable audit rules and imputations, known $\varphi_t\geq\varphi_{\min}>0$, binary outcomes, and a stable target $\E(Y_t\mid\mathcal F_{t-1})=\theta_0$. Then
\begin{equation}
\Prob\{\theta_0\in C_t\text{ for all }t\}\geq1-\alpha.
\label{eq:anytime}
\end{equation}
The guarantee is finite-sample and holds under adaptive auditing, repeated inspection and arbitrary stopping.
\end{theorem}

The stable target holds for one frozen agent on independently drawn tasks, not for clustered tasks, sampling without replacement from a small benchmark, agent updates, or mixed agents. For $K$ agents, level $\alpha/K$ per sequence gives simultaneous coverage. Across 200 streams monitored after every episode up to $t=3{,}000$, repeated nominal Wald intervals jointly cover only 0.535 (Wilson Monte-Carlo 95\% interval $[0.466,0.603]$), whereas the confidence sequence covers 0.995 ($[0.972,0.999]$). The price is width: at the final look, the confidence sequence was on average wider than the Wald interval.

% ---------------------------------------------------------------------------------------------
\section{\uppercase{Rank-Valid Leaderboards}}
\label{sec:ranks}

Let $K$ agents have abilities $\theta_a$ and competition ranks $r_a=1+\#\{b:\theta_b>\theta_a\}$, so tied agents share a rank.\textsuperscript{\ref{fn:ai}} Pairwise tests whose probability of any false directional declaration is at most $\alpha$ declare only the comparisons that the data support, and give the rank set
\begin{equation}
\begin{split}S_a=\big[&1+\#\{b:b\succ a\text{ declared}\},\\&K-\#\{b:a\succ b\text{ declared}\}\big].\end{split}
\label{eq:rankset}
\end{equation}
When abilities are only partially identified, with simultaneous outer intervals $D_a=[\ell_a,u_a]$, the identification-robust set is
\begin{equation}
S_a^{ID}=\big[1+\#\{b:\ell_b>u_a\},\,K-\#\{b:\ell_a>u_b\}\big],
\label{eq:rankID}
\end{equation}
which certifies only orderings that hold for every compatible ability vector. Both sets contain all true ranks with probability at least $1-\alpha$ (Section~\ref{sec:proofs}). With an alternating interval-center displacement of 0.025, 1,000 simulations give simultaneous coverage 1.000 ($[0.996,1.000]$) for \eqref{eq:rankID}, but only 0.037 ($[0.027,0.051]$) for rankings built from interval midpoints, whose exact recovery is 0/1,000.

Tasks from one software repository need not be independent: with 40 independent repositories, 10 tasks each and eight truly tied systems, treating tasks as independent gives a familywise false split rate of 0.188 ($[0.152,0.229]$) and simultaneous coverage 0.813. Repository-cluster Holm reduces false splits to 0.035 ($[0.021,0.058]$) and restores coverage to 0.965; a cluster wild-bootstrap Romano--Wolf stepdown keeps null coverage at 0.980.

\begin{algorithm}[htbp]
\caption{Transport-safe agent evaluation.}\label{alg:protocol}
\KwIn{agents, task population, judge outputs, optional calibration, audit budget, positivity floor $\varphi_{\min}$.}
Fix the target task population and keep task-level outcomes\;
If an audit budget exists, run a randomized pilot audit on a fixed subset, drawn independently with known probabilities $\geq\varphi_{\min}$\;
Diagnose operating curves, $J\neq0$ and support from calibration and pilot labels\;
\eIf{point identification is justified (Theorem~\ref{thm:onejudge}, or the audit)}{use the design-based route if audited, else the conditional route\;}{report sets (Proposition~\ref{prop:sharp}, or \eqref{eq:envelope} with $J_j>0$ and an externally justified $\Gamma$) and carry simultaneous intervals for them into \eqref{eq:rankID}; for a judge panel without a justified $\Gamma$, report the independence point and a sensitivity curve, and certify no ordering\;}
If audited, fit cross-fitted outcome models (pilot imputations use pilot labels only), learn \eqref{eq:kkt} from the pilot and draw the second wave independently among the episodes outside the pilot subset (Section~\ref{sec:audit})\;
If audited, for a stable target (each agent frozen, tasks drawn independently), update each agent's confidence sequence at level $\alpha/K$ after every episode, audited or not, with predictable $\widehat m_{t-1}\in[0,1]$ and audit rules, known $\varphi_t\geq\varphi_{\min}$, and a finite grid $\Lambda\subset(0,\varphi_{\min})$ fixed in advance (Theorem~\ref{thm:anytime})\;
Cluster comparisons at the task-generating unit with directional familywise control\;
\KwOut{simultaneous ability intervals and rank sets, not point scores alone.}
\end{algorithm}

\begin{figure*}[htbp]
\centering
\begin{tikzpicture}
\node[anchor=south west,inner sep=0] (img) at (0,0) {\includegraphics[width=\textwidth]{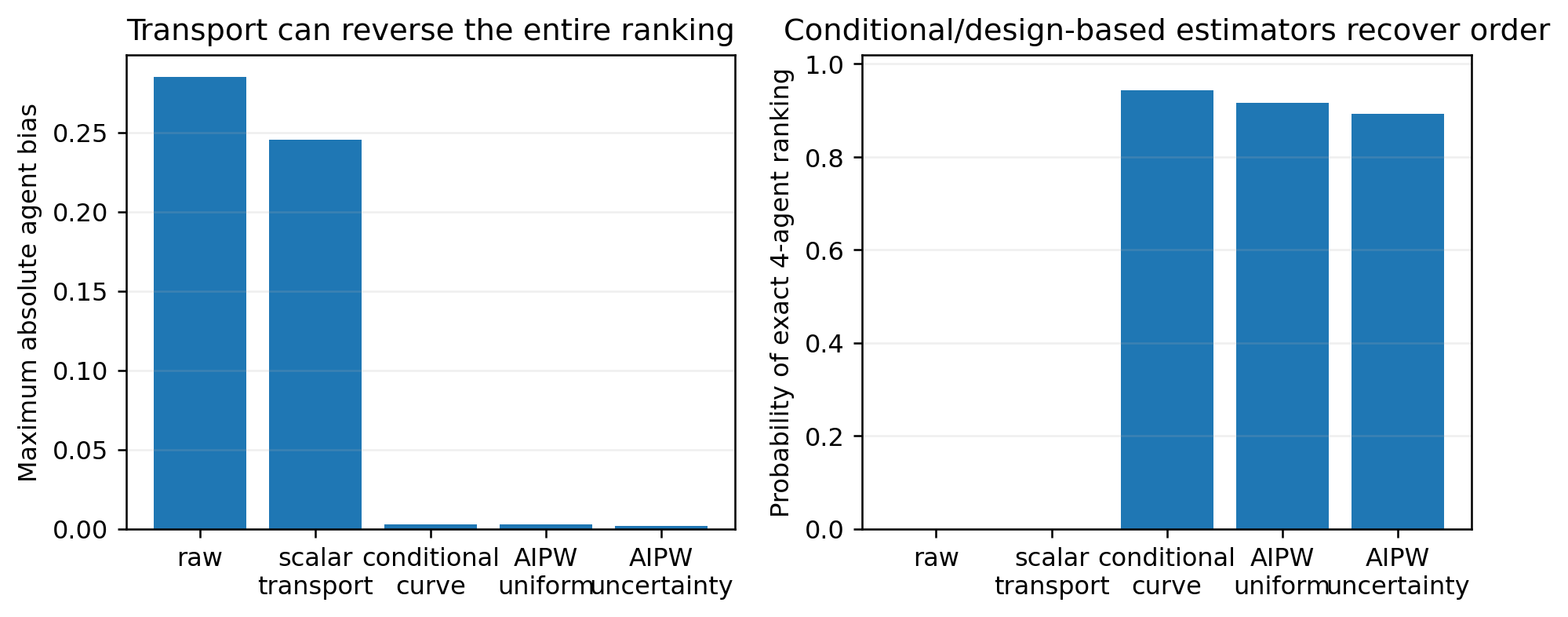}};
\begin{scope}[x={(img.south east)},y={(img.north west)},every node/.style={font=\sffamily\scriptsize,inner sep=1pt}]
  \fill[white] (0.08,0.0) rectangle (0.47,0.115);
  \fill[white] (0.555,0.0) rectangle (0.94,0.115);
  \fill[white] (0.075,0.915) rectangle (0.475,1.0);
  \fill[white] (0.495,0.915) rectangle (0.995,1.0);
  \node[anchor=north] at (0.274,0.995) {\small Scalar transport leaves a large bias};
  \node[anchor=north] at (0.744,0.995) {\small Conditional and audited estimators recover the order};
  \foreach \x/\l in {0.1276/raw, 0.2011/{scalar\\transport}, 0.2746/{conditional\\curve}, 0.3482/{AIPW\\uniform}, 0.4217/{AIPW\\uncertainty}}
    \node[align=center,anchor=north] at (\x,0.118) {\strut\l};
  \foreach \x/\l in {0.5973/raw, 0.6708/{scalar\\transport}, 0.7444/{conditional\\curve}, 0.8179/{AIPW\\uniform}, 0.8914/{AIPW\\uncertainty}}
    \node[align=center,anchor=north] at (\x,0.118) {\strut\l};
\end{scope}
\end{tikzpicture}
\caption{Four-agent benchmark, 250 replications, same episodes for all procedures. Left: maximum absolute agent bias. Right: probability of the exact ranking. ``Raw'' is the uncorrected judge rate; ``scalar transport'' is the oracle-calibrated scalar plug-in \cite{lee2025}; ``conditional curve'' is the conditional oracle of Table~\ref{tab:results}; ``AIPW uniform'' is the uniform-audit EIF/PPI baseline \cite{chen2026b}; ``AIPW uncertainty'' is design-aware AIPW.}
\label{fig:benchmark}
\end{figure*}

\begin{table*}[htbp]
\caption{Main results on transport and identification (other results in Sections~\ref{sec:audit}--\ref{sec:ranks}). Nominal mean pointwise coverage: 0.95. Brackets: Wilson 95\% intervals for proportions (agent-replications treated as independent); MC95 (Monte-Carlo 95\%) intervals for MAE; not task-population intervals.}
\label{tab:results}
\centering\footnotesize
\begin{tabular}{@{}>{\raggedright\arraybackslash}p{0.2\textwidth}>{\raggedright\arraybackslash}p{0.26\textwidth}>{\raggedright\arraybackslash}p{0.48\textwidth}@{}}
\toprule
Question & Compared procedures & Main result\\
\midrule
Does scalar calibration transport? & Scalar plug-in \cite{lee2025}; conditional oracle; uniform-audit EIF/PPI \cite{chen2026b}; design-aware AIPW & Scalar: MAE 0.168 (MC95 $[0.163,0.173]$), maximum bias 0.245, mean pointwise coverage 0.016 $[0.010,0.026]$, exact full ranking 0/250. Conditional oracle, uniform-audit AIPW, design-aware AIPW: MAE 0.018 $[0.017,0.019]$, 0.023 $[0.022,0.024]$, 0.021 $[0.020,0.022]$; coverage 0.939, 0.939, 0.950; maximum bias below 0.003 for both audited estimators; agent-wise bias intervals include zero.\\
Does rich $X$ make point correction data-hungry? & 48-cell plug-in with pooled fallback vs.\ 12\% randomized-audit AIPW & Effective support 0.209 to 0.805 for calibration $n=600$ to 4,800; plug-in MAE 0.188 to 0.077; AIPW 0.018--0.020.\\
Can judge panels replace human labels? & Three-judge moments under shared dependence vs.\ 10\% randomized audit & Label-free MAE rises to 0.0343; AIPW stays 0.0062--0.0096. Dependence sets cover the truth where the independence point fails.\\
\bottomrule
\end{tabular}
\end{table*}

% ---------------------------------------------------------------------------------------------
\section{\uppercase{Proof Sketches}}
\label{sec:proofs}

The proof sketches below were drafted with Claude \cite{anthropic2026} and checked by the authors.\textsuperscript{\ref{fn:ai}}

\textbf{Theorem~\ref{thm:transport}.} By blindness, $\Prob_a(\Yhat=1\mid Y=1)=\E_a\{s(X)\mid Y=1\}=s^\star_a$ and $\Prob_a(\Yhat=0\mid Y=0)=t^\star_a$, so $p_a=\Prob_a(\Yhat=1)=\theta_as^\star_a+(1-\theta_a)(1-t^\star_a)$. Since $J_0\neq0$, the continuous mapping theorem gives $\widehat\theta_a^{RG}\to(p_a+t_0-1)/J_0$. Subtracting $\theta_a=\theta_aJ_0/J_0$ from this limit leaves $\{\theta_a(s^\star_a-s_0)+(1-\theta_a)(t_0-t^\star_a)\}/J_0=B_a$. The covariance form follows from $\E_0\{w_{a,1}(X)\mid Y=1\}=1$. The reversal condition is $(\theta_a+B_a)-(\theta_b+B_b)<0$.

\textbf{Theorem~\ref{thm:onejudge}.} Let $q_a(x)=\Prob_a(Y=1\mid X=x)$. By blindness, $\mu_a=q_as+(1-q_a)(1-t)$, so $q_a=(\mu_a+t-1)/J$ on $\mathcal O$, where both curves are identified; since $\Prob_a(X\in\mathcal O)=1$, averaging over $\pi_a$ gives \eqref{eq:onejudge}. Non-identification follows from Proposition~\ref{prop:sharp}: on $\mathcal U_0$ every $q_a(x)\in[0,1]$ is compatible with the data, so the set has width at least $\Prob_a(X\in\mathcal U_0)>0$.

\textbf{Proposition~\ref{prop:sharp}.} $\theta_a=\nu_a+\E_a[\mathbf 1\{X\in\mathcal U\}q_a(X)]$ and $q_a\in[0,1]$, which gives validity. On $\mathcal U_0$, any $q_a(x)\in[0,1]$ is compatible with the observed $\mu_a(x)$: take $s(x)=\mu_a(x)$ and $t(x)=1-\mu_a(x)$; when both curves are identified with $J(x)=0$, this is the true pair, since then $\mu_a(x)=s(x)$. Hence $q_a\equiv0$ and $q_a\equiv1$ on $\mathcal U$ attain both endpoints when $\Prob_a(X\in\mathcal U_1)=0$, and mixtures attain every value between them. On $\mathcal U_1$ the known curve can restrict $q_a(x)$ (with $t$ known, $q_a=(\mu_a+t-1)/(s+t-1)$ for free $s$), so the set is then outer. Pooling agents can also shrink it: by blindness, $\mu_b(x)=0$ and $\mu_c(x)=1$ at $x\in\mathcal U_0$ force $q_a(x)\in\{\mu_a(x),1-\mu_a(x)\}$. With bounds, replace $0$ and $1$ by $\ell$ and $u$.

\textbf{Theorem~\ref{thm:three}.} Fix $x$. Conditional independence gives $\Yhat_j-\mu_j=J_j(Y-\theta)+\varepsilon_j$, with $\E(\varepsilon_j\mid Y)=0$ and $\varepsilon_1,\varepsilon_2,\varepsilon_3$ independent given $Y$. Terms containing any $\varepsilon_j$ have zero mean, so $c_{jk}=J_jJ_k\,\mathrm{Var}(Y)=vJ_jJ_k$ and $T=J_1J_2J_3\,\E(Y-\theta)^3=v(1-2\theta)J_1J_2J_3$. With $r=T^2/(c_{12}c_{13}c_{23})=(1-2\theta)^2/v=(1-4v)/v$, we get $v=1/(r+4)$; since $v>0$ and $J_j>0$, the sign of $T$ places $\theta$ on the correct side of $1/2$ ($T=0$ gives $\theta=1/2$). Then $J_1^2=c_{12}c_{13}/(vc_{23})$, similarly for $J_2,J_3$, and $\mu_j=1-t_j+\theta J_j$ gives $t_j$ and $s_j$. Two judges give three moments $(\mu_1,\mu_2,c_{12})$ for five unknowns.

\textbf{Envelope set.} Assume $J_j>0$ as in Theorem~\ref{thm:three}. The true $(\theta,J)$ is feasible; if its deviation is within $\Gamma$, then $\theta\in I_\Gamma$. The converse can fail: compatibility with the full judge law $p$ also needs some $g:\{0,1\}^3\to[0,1]$ with $\sum_hp(h)g(h)=\theta$ and $\sum_hp(h)g(h)(h_j-\mu_j)=\theta(1-\theta)J_j$.

\textbf{AIPW unbiasedness and allocation.} Given $(V_{1:n},Y_{1:n})$ and the other folds, $\E\{R_i/\varphi(V_i)\}=1$, so each summand of \eqref{eq:aipw} has expectation $Y_i$: the imputation error cancels for any $\widehat m$ not using episode $i$'s label. For \eqref{eq:kkt}, the asymptotic variance is $\mathrm{Var}\{m(V)\}+\E\{\sigma^2(V)/\varphi(V)\}$; by convexity of $\sigma^2/\varphi$, the Lagrangian under a fixed $\E\{\varphi\}$ gives $\lambda\sigma$ clipped to $[\varphi_{\min},1]$; $\lambda^\star$ exists because $\lambda\mapsto\E\{\varphi^\star(V)\}$ is continuous and nondecreasing, equals $\varphi_{\min}$ at $\lambda=0$ and tends to $\varphi_{\min}+(1-\varphi_{\min})\Prob\{\sigma(V)>0\}$. In two waves, conditioning also on the pilot wave fixes the second-wave $\varphi$, so the same argument applies to second-wave summands; pilot summands are covered by the first argument, since their imputations use only pilot audits. With finitely many strata, consistent pilot estimates $\widehat\sigma_s$ and continuity of $\sigma\mapsto\varphi^\star$ ($\varphi^\star$ is unique even where $\lambda^\star$ is not) give convergence to $\varphi^\star$.

\textbf{Theorem~\ref{thm:anytime}.} Conditioning first on $(\mathcal F_{t-1},V_t,Y_t)$ gives $\E\{W_t(\theta_0)\mid\mathcal F_{t-1},V_t,Y_t\}=Y_t-\theta_0$, so $\E\{W_t(\theta_0)\mid\mathcal F_{t-1}\}=0$ by the stable-target condition. Since $\widehat m_{t-1},Y_t,\theta\in[0,1]$, $|W_t|\leq1/\varphi_t\leq1/\varphi_{\min}$; hence $1\pm\lambda W_t>0$ for $\lambda<\varphi_{\min}$. Each product in \eqref{eq:eprocess} is a nonnegative martingale starting at 1, and so is their average $E_t(\theta_0)$. Ville's inequality \cite{howard2021} gives $\Prob\{\exists t:E_t(\theta_0)\geq1/\alpha\}\leq\alpha$, which is \eqref{eq:anytime}; it holds at any stopping time.

\textbf{Rank sets.} If no false direction is declared (a strict order between tied agents counts as false), $1+\#\{b\succ a\}\leq r_a\leq K-\#\{a\succ b\}$; the familywise guarantee gives probability at least $1-\alpha$. For \eqref{eq:rankID}, $\ell_b>u_a$ implies $\theta_b>\theta_a$ for every ability vector in the intervals, so $\Prob(r_a\in S_a^{ID}\ \forall a)\geq\Prob(\theta_b\in D_b\ \forall b)\geq1-\alpha$.

% ---------------------------------------------------------------------------------------------
\section{\uppercase{Evidence and Example}}
\label{sec:evidence}

Table~\ref{tab:results} summarizes the main tests.\textsuperscript{\ref{fn:ai}} \emph{Newly executed}: all simulation studies, including the four-agent benchmark (\figref{fig:benchmark}). \emph{Source-preserved}: the AgentRewardBench operating curves and a historical rank audit of 134 SWE-bench Verified systems (not publicly released), which need an immutable snapshot before they can be called independently rerun. 

\textbf{Worked example: quality and cost.} We use two separate simulations. \emph{Rich trajectories (48 cells).} With 600 calibration labels, cells with enough labels hold on average only 20.9\% of the target mass, and the per-cell plug-in (pooled fallback) has mean absolute error 0.188; with 4,800 labels it is still 0.077. Labeling a uniform random 12\% of the target episodes and using AIPW gives 0.018--0.020 at every size; compared with labeling everything by hand, only 12\% of the target episodes (in expectation) need a human label. Label counts are not compared, as the number of target episodes is not reported here. \emph{Four agents.} The scalar shortcut never recovers the true ranking in 250 replications, whereas the audited estimators recover it with high probability (\figref{fig:benchmark}, right) and have mean pointwise coverage close to the nominal 0.95 (0.939--0.950). Each e-process update costs $O(|\Lambda|)$ per grid value of $\theta$, plus refitting $\widehat m$.

% ---------------------------------------------------------------------------------------------
\section{\uppercase{Discussion}}
\label{sec:discussion}

\textbf{Assumptions.} Adding judges does not repair missing calibration support except under a model:\textsuperscript{\ref{fn:ai}} three conditionally independent judges with $J_j>0$ and $0<\Prob_a(Y=1\mid X)<1$ give a model-based point, and, still with $J_j>0$, a justified external envelope gives only a set. A randomized audit needs no correct judge model, but needs independently drawn audits with known inclusion probabilities, positivity, imputations that never use the episode's own label, and valid human labels; unbiasedness for $\theta_a$ also assumes independently drawn episodes, and monitoring needs predictable rules and a stable target.

\textbf{Limitations.} The main results are semi-synthetic; real agent data with live LLM judges is still needed. The public-data analyses are not rerun; their judge settings (temperature, top-p, seed) are available only to the extent documented upstream. A paired intervention on trajectory renderings is preregistered and powered but not run, so no causal effect of length is claimed. Expert labels are treated as the reference. The rich-$X$ study is discretized; continuous high-dimensional nuisance learning and ordinal or multi-objective quality remain open. Envelope sets need a defensible external envelope, and Monte-Carlo intervals measure simulation precision, not external validity. Rank inference targets a task superpopulation; a fixed benchmark's descriptive scores are not random once all outcomes are observed.

% ---------------------------------------------------------------------------------------------
\section{\uppercase{Conclusion}}
\label{sec:conclusion}

A judge's error rates can depend on the trajectories it reads, so they cannot in general be carried from one agent to another.\textsuperscript{\ref{fn:ai}} We gave the exact limit of this error and a protocol (Algorithm~\ref{alg:protocol}) that avoids it under stated design assumptions: correct only where calibration covers the target and the judge is informative, otherwise report a set; audit a random sample with known probabilities; monitor with a confidence sequence; and report rank sets rather than a single order. The next step is to run this protocol on real agent evaluations with live LLM judges, starting with the preregistered test of trajectory length.

\section*{\uppercase{Acknowledgements}}

\begin{itemize}\setlength{\itemsep}{0pt}
\item \textbf{AI Tools:} ChatGPT \cite{openai2026} assisted the submitted version; Claude \cite{anthropic2026} drafted or edited all sections of this version, including the proofs. The authors checked every source, statement and number; no data or results were AI-generated.
\end{itemize}

\bibliographystyle{apalike}
{\small
\bibliography{references_v2}}

\end{document}